\def\year{2021}\relax
\documentclass[letterpaper]{article} 
\usepackage{aaai21}  
\usepackage{times}  
\usepackage{helvet} 
\usepackage{courier}  
\usepackage[hyphens]{url}  
\usepackage{graphicx} 
\usepackage{natbib}  
\usepackage{caption} 
\usepackage{multirow}
\usepackage{booktabs}
\usepackage{amssymb}
\title{A Hierarchical Transformer with Speaker Modeling for Emotion Recognition in Conversation}
\author{
    Jiangnan Li \textsuperscript{\rm 1,2}, 
    Zheng Lin \textsuperscript{\rm 1}, 
    Peng Fu \textsuperscript{\rm 1}, 
    Qingyi Si \textsuperscript{\rm 1,2}, 
    Weiping Wang \textsuperscript{\rm 1}
    \\
}
\affiliations{
    \textsuperscript{\rm 1}Institute of Information Engineering, Chinese Academy of Sciences, Beijing, China \\
    \textsuperscript{\rm 2}School of Cyber Security, University of Chinese Academy of Sciences, Beijing, China \\
    \{lijiangnan, linzheng, fupeng, siqingyi, wangweiping\}@iie.ac.cn

}

\begin{document}

\maketitle

\begin{abstract}
Emotion Recognition in Conversation (ERC) is a more challenging task than conventional text emotion recognition. It can be regarded as a personalized and interactive emotion recognition task, which is supposed to consider not only the semantic information of text but also the influences from speakers. The current method models speakers’ interactions by building a relation between every two speakers. However, this fine-grained but complicated modeling is computationally expensive, hard to extend, and can only consider local context. To address this problem, we simplify the complicated modeling to a binary version: \emph{Intra-Speaker} and \emph{Inter-Speaker} dependencies, without identifying every unique speaker for the targeted speaker. To better achieve the simplified interaction modeling of speakers in Transformer, which shows excellent ability to settle long-distance dependency, we design three types of masks and respectively utilize them in three independent Transformer blocks. The designed masks respectively model the conventional context modeling, \emph{Intra-Speaker} dependency, and \emph{Inter-Speaker} dependency. Furthermore, different speaker-aware information extracted by Transformer blocks diversely contributes to the prediction, and therefore we utilize the attention mechanism to automatically weight them. Experiments on two ERC datasets indicate that our model is efficacious to achieve better performance.

\end{abstract}

\section{Introduction}


Nowadays, intelligent machines to precisely capture speakers' emotions in conversations are gaining popularity, thus driving the development of Emotion Recognition in Conversation (ERC). ERC is a task to predict the emotion of the current utterance expressed by a specific speaker according to the context \citep{Survey}, which is more challenging than the conventional emotion recognition only considering semantic information of an independent utterance. 


\begin{figure}
  \centering
  \includegraphics[width=0.45 \textwidth]{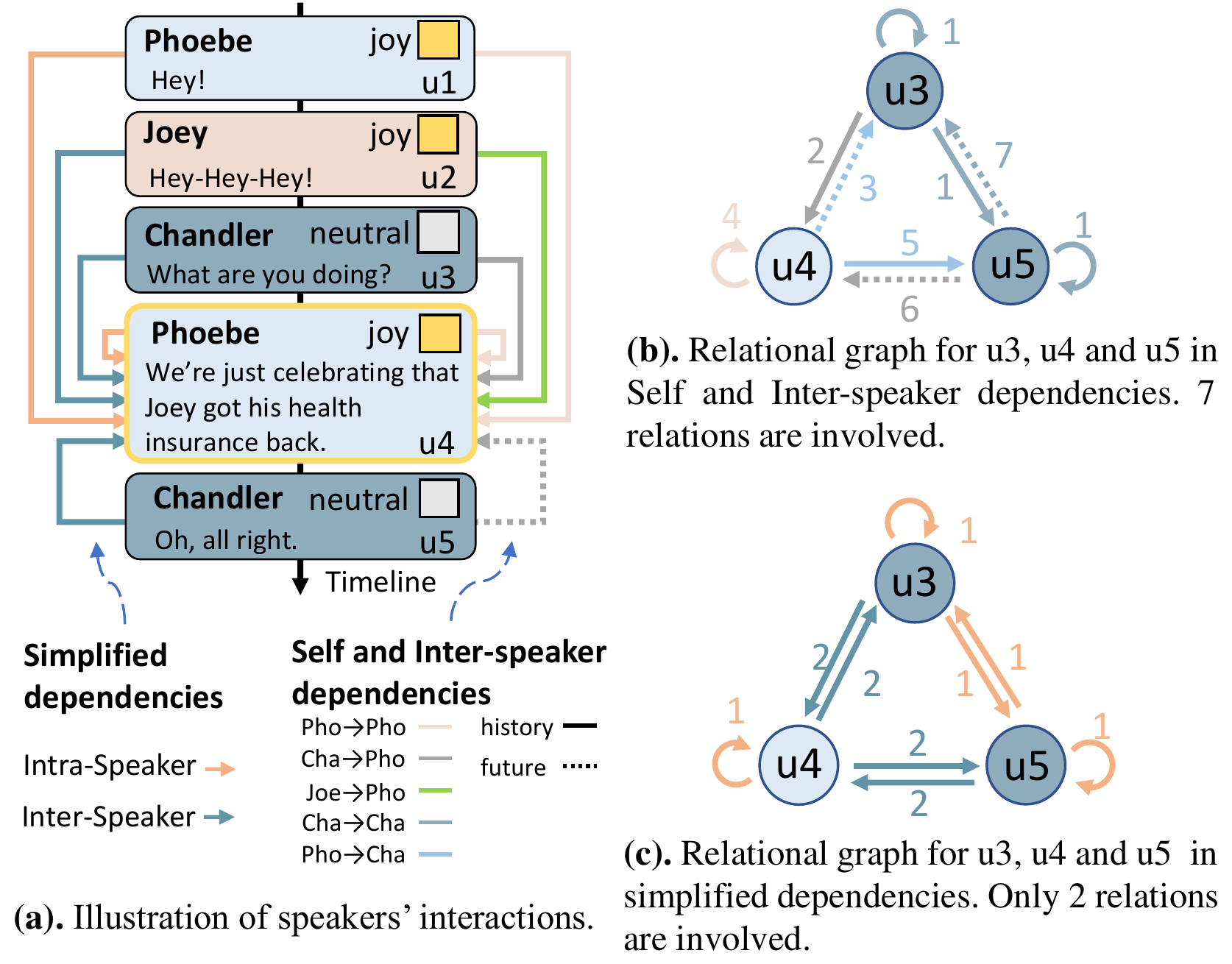}
  \caption{(a) illustrates a conversation clip of 3 speakers and the utterance in yellow frame is selected as a targeted utterance; (b) and (c) illustrate the relation graphs of u3, u4, u5 in different dependencies. }
  \label{fig: eg}
\end{figure}




To precisely predict the emotion of a targeted utterance, both the semantic information of the utterance and the information provided by utterances in the context are critical. Nowadays, a number of works \citep{ICON,CMN,DialogueRNN,DialogueGCN} demonstrate that the interactions between speakers can facilitate extracting information from contextual utterances. We denote this kind of information with modeling speakers' interactions as speaker-aware contextual information. To capture speaker-aware contextual information, the state-of-the-art model DialogueGCN \citep{DialogueGCN} introduces Self and Inter-Speaker dependencies, which capture the influences from different speakers. As illustrated in Fig. \ref{fig: eg} (a), Self and Inter-Speaker dependencies establish a specific relation between every two speakers and construct a fully connected relational graph. And then a Relational Graph Convolutional Network (RGCN) \citep{RGCN} is applied to process such a graph. 

Although DialogueGCN can achieve excellent performance with Self and Inter-Speaker dependencies, this speaker modeling is easy to be complicated with the number of speakers increasing. As shown in Fig. \ref{fig: eg} (b), for a conversation clip with two speakers, the considered relations reach to 7. The number can drastically increase with more speakers involved. Thus this complicated speaker modeling is hard to deal with the condition that the number of speakers dynamically changes, and not flexible to be deployed in other models. In addition, RGCN processing the fully connected graph with multiple relations requires tremendous consumption of computation. This limitation leads to DialogueGCN only considering the local context in a conversation \citep{DialogueGCN}. Therefore, it is appealing to introduce a simple and general speaker modeling, which is easy to extend in all scenes and realize in other models so that long-distance context can be available.

To address the above problem, we propose a TRansforMer with Speaker Modeling (TRMSM). First, we simplify the Self and Inter-speaker dependencies to a binary version, which only contains two relations: \emph{Intra-Speaker} dependency and \emph{Inter-Speaker} dependency. As illustrated in Fig. \ref{fig: eg}, for the speaker of the targeted utterance, \emph{Intra-Speaker} dependency focuses on the influence from the same speaker, and \emph{Inter-Speaker} dependency treats other speakers as a whole group instead of building a relation between every two speakers. In this way, our simplified modeling can be easy to extend in other models and deal with the scene with the dynamical number of speakers without introducing new relations between speakers. 

Furthermore, with the ability to settle long-distance dependency, Transformer \citep{Transformer} achieves excellent performance among a great number of Natural Language Processing (NLP) problems. To better model the long-distance contextual utterances in a conversation, we utilize a hierarchical Transformer with two levels: sentence level and dialogue level. In the sentence level, BERT \citep{BERT} encodes the semantic representation for a targeted utterance, and in the dialogue level, Transformer is used to capture the information from contextual utterances. To better model our simplified dependencies in the dialogue-level Transformer, we design three masks: Conventional Mask for conventional context modeling, Intra-Speaker Mask for \emph{Intra-Speaker} dependency, and Inter-Speaker Mask for \emph{Inter-Speaker} dependency. To realize the functions of masks, we deploy three independent Transformer blocks in dialogue level and the designed masks are respectively used in these Transformer blocks. With different speaker-aware contextual information extracted by these Transformer blocks, whose contributions to the final prediction are diverse, we utilize the attention mechanism to automatically weight and fuse them. Besides, we also apply two other simple fusing methods: Add and Concatenation to demonstrate the advancement of the attention. 

Specifically, our contributions are concluded as follows:

\begin{itemize}
\item We simplify Self and Inter-speaker dependencies to a binary version, so that the speaker interaction modeling can be extended in hierarchical Transformer and the long-distance context can be considered. 
\item We design three types of masks to achieve speakers' interactions modeling in Transformer and utilize the attention mechanism to automatically pick up the important speaker-aware contextual information. 
\item We conduct experiments on two ERC datasets: IEMOCAP and MELD. Our method achieves state-of-the-art performance on both datasets on average. 
\end{itemize}

\section{Related Work}

Two aspects are strongly related to our work: Emotion recognition in conversation and Utilization of mask in Transformer. 

\textbf{Emotion recognition in conversation} \quad Hierarchical structure based on RNN \citep{AGHMN,HiGRUs} or Transformer \citep{KET,HiTrans} is leveraged in ERC to capture contextual information. Except contextual information, speaker information is proven to be important to ERC. Speakers can be regarded as objects related to utterances or additional information for utterances. As objects, speakers are involved in the graph of conversation as nodes to interact with utterances \citep{ConvGCN}. As additional information, speaker information is modeled via utterances. Specifically, \citet{CMN,ICON} employ GRUs and Memory Network (Memnet) \citep{MN} to model speakers' interactions in the dyadic conversation, which is difficult to extend to multi-speaker conditions. Therefore, \citet{DialogueRNN} generalize speakers as parties, track them by GRU, and utilize attention mechanism to gather interactive information in multi-speaker conversations. Even so, \citet{DialogueGCN} argue that \citet{DialogueRNN} ignored the influences from other speakers and propose Self and Inter-Speaker dependencies to formalize interactions within and between speakers. However, the complicated modeling of speakers' interactions is difficult to apply in other models, thus requiring a simplified version. 

\textbf{Utilization of mask in Transformer} \quad Masks in Transformer are utilized to mask the unattended elements in self-attention. Recently, masks are well-designed and leveraged in language modeling \citep{UniLM,BERT,GPT} and conversation structure modeling \citep{StrucureTrans}. Masks are flexible and convenient to be implemented and we choose them to model the interactions of speakers in Transformer. 



\section{Methodology}

\begin{figure*}
    \centering
    \includegraphics[width=1\textwidth]{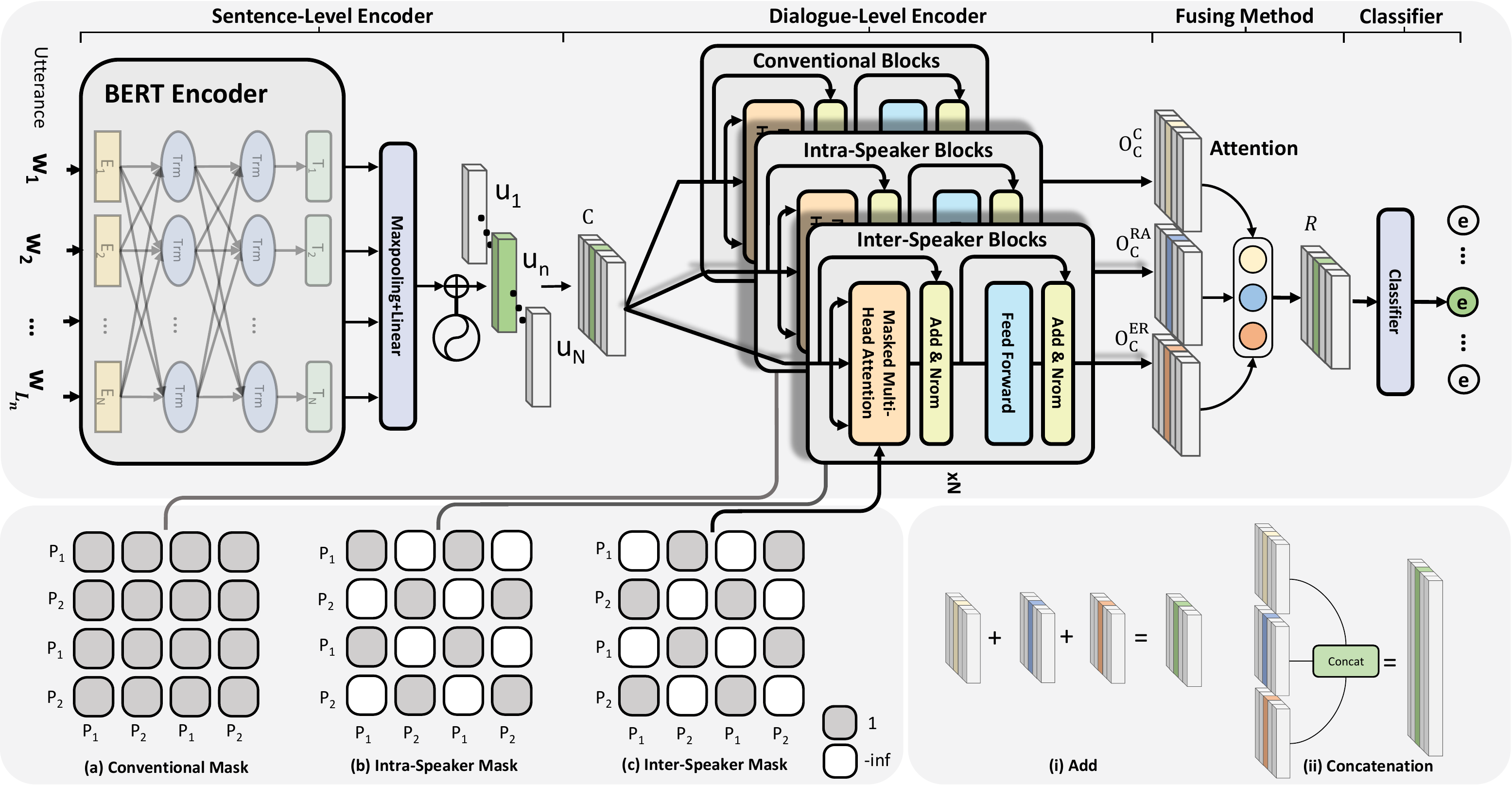}
    \caption{The structure of our proposed model, which is based on Transformer structure. Our proposed masks are utilized in the Multi-Head Attention of Dialogue-Level Encoder and are illustrated for 3 types: (a)conventional, (b)Intra-Speaker and (c)Inter-Speaker masks. The fusing methods include Attention, (i)Add, and (ii)Concatenation.  }
    \label{fig: model}
\end{figure*}

In this section, we will elaborate on the task definition and the structure of TRMSM which is illustrated in Fig. \ref{fig: model}. Our model contains 4 parts: Sentence-Level Encoder, Dialogue-Level Encoder, Fusing Method, and Classifier. 

\subsection{Task Definition}

ERC task includes $K$ emotions, whose set is $E=\{emo_1, emo_2, ..., emo_K\}$. Given a conversation $C=[u_1, u_2, ..., u_N]$ containing \textit{N} textual utterances, each utterance $u_n=[w_1, w_2, ..., w_{L_n}]$ within is sequentially formed by $L_n$ words. Particularly, $M$ speakers, whose set is $SPK=\{spk_1,spk_2,...,spk_M\}$, participate in the conversation. For each utterance $u_n$, a emotion label $e_n \in E$ and a speaker annotation $p_n \in SPK$ are assigned. ERC task aims to predict the emotion of every utterance in $C$ with the information provided above. 


\subsection{Sentence-Level Encoder}

To encode a more informative and context-aware representation of a single utterance based on Transformer, we utilize a BERT encoder. Limited by the max length of a sequence supported by BERT, we cannot input the concatenated sequence of all utterances in a conversation, whose length frequently exceeds 768 in cases of long conversations, to capture the global contextual information. Therefore, BERT is solely used to encode the sentence-level context in a single utterance. An utterance $u_n=[ w_1, w_2,...,w_{L_n}]$ is fed into BERT to obtain the contextualized representation of words:
\begin{equation}
    W=BERT(w_1, w_2,...,w_{L_n}) \label{eq: 1}
\end{equation}
where $W \in {\mathbb{R}^{L_n \times d_w}}$ is the output of the top layer of BERT and $d_w$ is the dimension of the word representation. To obtain an utterance representation for $u_n$, a max-pooling operation followed by a projection is deployed: 
\begin{equation}
    u_n = Linear(Maxpooling(W)) \label{eq: 2}
\end{equation}
where $u_n \in \mathbb{R}^{d_u}$ represents the utterance and $d_u$ is the dimension of utterance representation. By processing every utterance in a conversation, we finally obtain the representation matrix $C \in \mathbb{R}^{{N} \times {d_u}}$.

\subsection{Dialogue-Level Encoder} \label{sec: 3.2.2}


In the dialogue level, we utilize three transformer blocks: Conventional Blocks for conventional context modeling, Intra-Speaker Blocks for \emph{Intra-Speaker} dependency, and Inter-Speaker Blocks for \emph{Inter-Speaker} dependency. Due to the same structures of all Transformer blocks, we simply introduce the general process of the first layer of Transformer blocks. 

Given the conversation matrix $C$ processed by the sentence-level encoder, to avoid the absence of positional information in $C$, an Absolute Positional Embedding is added to every representation in $C$:
\begin{equation}
    C = C + PE(0:N) \label{eq: 3}
\end{equation}
where $PE(0:N)$ is in the same dimension as C. 

Self-attention intuitively provides an interactive pattern for contextual modeling of conversations. Taking advantage of the mechanism of self-attention, the targeted utterances can be parallelly processed. Therefore, the targeted utterances are regarded as a query matrix, and the contextual utterances act as a key matrix, so that every utterance simultaneously assesses how much information shall be obtained from every contextual utterance. In this way, $C$ is projected to query matrix $Q \in \mathbb{R}^{N\times d_a}$, key matrix $K\in \mathbb{R}^{N\times d_a}$, and value matrix $V\in \mathbb{R}^{N\times d_a}$ by linear projections without bias: $[Q;K;V] = Linear([C;C;C])$ where $[ ]$ is the concatenating operation. Self-attention is calculated by: 
\begin{equation}
    A(Q, K, V, M)=softmax(\frac{(Q{K^T})*M}{\sqrt{d_a}})V  \label{eq: 4}
\end{equation}
where $*$ denotes element-wise multiplication; $M \in \mathbb{R}^{N \times N}$ is the utilized mask which is a square matrix whose non-infinite elements equal 1. We will introduce different masks used by diverse blocks later. Transformer hires multiple self-attention (Multi-Head Attention, MHA) to model different aspects of information. And then the outputs of all heads are concatenated and projected to $O$ with the same size of $C$. After the Attention module, a Position-wise Feed-Forward Network (FFN) module is deployed to produce output $F \in \mathbb{R}^{N\times d_u}$. MHA and FFN are both residually connected. Therefore the output $O_{C}^{1}$ of the first layer of Transformer is:
\begin{eqnarray}
& {A^{'}}=LayerNorm(O+C), \\
& F=max(0, {A^{'}}{W_{1}}+{b_{1}}){W_{2}}+{b_{2}}, \\
& O_{C}^{1}=LayerNorm(F+A^{'}).
\end{eqnarray}

$O_{C}^{1}$ acts as the input of the second transformer layer, and by this analogy, we obtain the final output $O_C \in \mathbb{R}^{N\times d_u}$ after multiple layers. Therefore, the outputs of our 3 blocks can be denoted as: $O_{C}^{C}$ for Conventional Block, $O_{C}^{RA}$ for Intra-Speaker Block, and $O_{C}^{ER}$ for Inter-Speaker Block. Due to the limited space in this paper, more details about Transformer can be reviewed in \citet{Transformer}. 

Masks can prompt Transformer blocks to realize their different functions, and we introduce how to form these 3 masks: 

\textbf{Conventional Mask} sets all the elements of itself to 1, which means that every targeted utterance can get access to all the contextual utterances. Conventional Mask is applied in the multi-head attention of Conventional Blocks and is illustrated in Fig. \ref{fig: model} (a). We annotate Conventional Mask as $M_{C}$. 

\textbf{Intra-Speaker Mask} only considers those contextual utterances tagged with $p_n$, which is the speaker tag of the targeted utterance. Therefore, based on $M_{C}$, Intra-Speaker Mask $M_{RA}$ sets positions representing other speakers to \textsc{-inf}. Intra-Speaker Mask is illustrated in Fig. \ref{fig: model} (b). 

\textbf{Inter-Speaker Mask} regards other speakers different from the one of the targeted utterance as one unit due to our simplification. Therefore, based on $M_{C}$, Inter-Speaker Mask $M_{ER}$ sets positions whose speaker is the same as the speaker tag of the targeted utterance to \textsc{-INF}. Inter-Speaker Mask is illustrated in Fig. \ref{fig: model} (c). 

\subsection{Fusing Method}
As blocks produce different outputs that carry various speaker-aware contextual information, we utilize 3 simple methods to fuse the information. 

\textbf{Add}\quad As illustrated in Fig. \ref{fig: model} (i), Add equally regards the contributions of all outputs of blocks. Therefore, the fusing representation is: 
\begin{equation}
    R=O_{C}^{C}+O_{C}^{RA}+O_{C}^{ER}
\end{equation}

\textbf{Concatenation}\quad Concatenation (illustrated in Fig. \ref{fig: model} (ii)) is also a simple but effective method to combine different information. Different from Add operation, Concatenation can implicitly choose the information which is important for the final prediction due to the following linear projection of classifier. Therefore, the fusing representation $R \in \mathbb{R}^{N\times 3d_u}$ is:
\begin{equation}
    R=Concat(O_{C}^{C}, O_{C}^{RA}, O_{C}^{ER}, dim=1)
\end{equation}

\textbf{Attention}\quad As the contributions of different speaker parties are diversely weighted, it is feasible that the model automatically chooses the more important information. Therefore, we utilize the widely used attention \citep{Fusion} to achieve this goal. Attention mechanism takes 3 block outputting representations as inputs and produces an attention score for each representation. For simplicity, we take representations ${O_{C}^{C}}_i \in \mathbb{R}^{1\times d_u}$, ${O_{C}^{RA}}_i \in \mathbb{R}^{1\times d_u}$, and ${O_{C}^{ER}}_i \in \mathbb{R}^{1 \times d_u}$ of utterance $i$ as an example. Therefore, the attention score and fusing representation are computed as:

\begin{eqnarray}
& O_i=Concat({O_{C}^{C}}_i,{O_{C}^{RA}}_i,{O_{C}^{ER}}_i, dim=0), \\
& \alpha =softmax(w_{F}{O_{i}}^T), \\
& R_{i}=\alpha O_{i}.
\end{eqnarray}
where $O_i \in \mathbb{R}^{3\times d_u}$ is the concatenated representations, $\alpha \in \mathbb{R}^{1\times 3}$ is the attention score, $w_{F} \in \mathbb{R}^{1\times d_u}$ is a trainable parameter, and $R_{i} \in \mathbb{R}^{1\times d_{u}}$ is the fusing representation. Finally, all fusing representations of utterances are concatenated as $R \in \mathbb{R}^{N\times d_{u}}$. 

\begin{table}[]
\centering
\scalebox{0.7}{
\begin{tabular}{@{}c|cc|c|c|c|c|c|c@{}}
\toprule
\multirow{2}{*}{Dataset} & \multicolumn{3}{c|}{\begin{tabular}[c]{@{}c@{}}Num. of\\ dialogues\end{tabular}} & \multicolumn{3}{c|}{\begin{tabular}[c]{@{}c@{}}Num. of\\ utterances\end{tabular}} & \multicolumn{2}{c}{\begin{tabular}[c]{@{}c@{}}Avg. length\\ of dialogue\end{tabular}} \\ \cline{2-9} 
                         & \multicolumn{1}{c|}{train}               & dev               & test              & train                      & dev                       & test                     & train/dev                                    & test                                    \\ \hline \hline
IEMOCAP                  & \multicolumn{2}{c|}{120}                                     & 31                & \multicolumn{2}{c|}{5810}                              & 1623                     & 48                                         & 52                                    \\ \hline
MELD                     & \multicolumn{1}{c|}{1039}                                     & 114               & 280               & 9989                       & 1109                      & 2610                     & 10                                          & 9                                     \\ \bottomrule
\end{tabular}
}
\caption{Statistics about IEMOCAP and MELD. }
\label{tab:dataset}
\end{table}

\subsection{Classifier}

With the sentence-level and dialogue-level contextual information fully modeled by encoders, the dialogue-level output is fed to a classifier which predicts the final emotion distributions: 
\begin{equation}
    \hat{Y}=softmax(RW_{clf}+b_{clf}) \label{eq:13}
\end{equation}
where $W_{clf}\in \mathbb{R}^{d_u\times K}$($W_{clf}\in \mathbb{R}^{3d_u\times K}$ for Concatenation), $b_{clf}\in \mathbb{R}^{K}$ and $\hat{Y}$ is the matrix of emotion distributions of all utterances in conversation $C$. The model is trained by a cross-entropy loss function, which is calculated as:
\begin{equation}
    L=-\frac{1}{\sum_{l=1}^{T} N_l}\sum_{l=1}^{T} \sum_{i=1}^{N_l} \sum_{e=1}^{K}y_i^elog(\hat{Y}_{i}^e)
\end{equation}
where $y_i$ is the one-hot vector denoting the emotion label of utterance $i$ in a conversation, $e$ denotes the dimension of each emotion, $N_l$ denotes the length of \emph{l}-th conversation, and $T$ denotes the number of conversations in a dataset. 

\begin{table*}
\centering
\scalebox{0.78}{
\begin{tabular}{@{}l|cccccccccccccc@{}}
\toprule
\multirow{3}{*}{Methods} & \multicolumn{14}{c}{IEMOCAP}                                                                                                                                                                                                                                                                                                                                                                                                                           \\ \cline{2-15} 
                         & \multicolumn{2}{c|}{Happy}                                      & \multicolumn{2}{c|}{Sad}                                        & \multicolumn{2}{c|}{Neutral}                                    & \multicolumn{2}{c|}{Angry}                                      & \multicolumn{2}{c|}{Excited}                                    & \multicolumn{2}{c|}{Frustrated}                                 & \multicolumn{2}{c}{Average}                \\ \cline{2-15} 
                         & \multicolumn{1}{c|}{Acc.} & \multicolumn{1}{c|}{F1}             & \multicolumn{1}{c|}{Acc.} & \multicolumn{1}{c|}{F1}             & \multicolumn{1}{c|}{Acc.} & \multicolumn{1}{c|}{F1}             & \multicolumn{1}{c|}{Acc.} & \multicolumn{1}{c|}{F1}             & \multicolumn{1}{c|}{Acc.} & \multicolumn{1}{c|}{F1}             & \multicolumn{1}{c|}{Acc.} & \multicolumn{1}{c|}{F1}             & \multicolumn{1}{c|}{Acc.} & wF1            \\ \hline\hline
Memnet\cite{MN}\dag                   & 25.72                     & \multicolumn{1}{c|}{33.53}          & 55.53                     & \multicolumn{1}{c|}{61.77}          & 58.12                     & \multicolumn{1}{c|}{52.84}          & 59.32                     & \multicolumn{1}{c|}{55.39}          & 51.50                     & \multicolumn{1}{c|}{58.30}          & 67.20                     & \multicolumn{1}{c|}{59.00}          & 55.72                     & 55.10          \\
CMN\cite{CMN}\dag                      & 25.00                     & \multicolumn{1}{c|}{30.38}          & 55.92                     & \multicolumn{1}{c|}{62.41}          & 52.86                     & \multicolumn{1}{c|}{52.39}          & 61.76                     & \multicolumn{1}{c|}{59.83}          & 55.52                     & \multicolumn{1}{c|}{60.25}          & 71.13                     & \multicolumn{1}{c|}{60.69}          & 56.56                     & 56.13          \\
DialogueRNN\cite{DialogueRNN}              & 25.69                     & \multicolumn{1}{c|}{33.18}          & 75.10                     & \multicolumn{1}{c|}{\textbf{78.80}} & 58.59                     & \multicolumn{1}{c|}{59.21}          & 64.71                     & \multicolumn{1}{c|}{\textbf{65.28}} & 80.27                     & \multicolumn{1}{c|}{71.86}          & 61.15                     & \multicolumn{1}{c|}{58.91}          & 63.40                     & 62.75          \\
KET\cite{KET}                      & -                         & \multicolumn{1}{c|}{-}              & -                         & \multicolumn{1}{c|}{-}              & -                         & \multicolumn{1}{c|}{-}              & -                         & \multicolumn{1}{c|}{-}              & -                         & \multicolumn{1}{c|}{-}              & -                         & \multicolumn{1}{c|}{-}              & -                         & 59.56          \\
AGHMN\cite{AGHMN}                    & 48.30                     & \multicolumn{1}{c|}{\textbf{52.10}}          & 68.30                     & \multicolumn{1}{c|}{73.30}          & 61.60                     & \multicolumn{1}{c|}{58.40}          & 57.50                     & \multicolumn{1}{c|}{61.90}          & 68.10                     & \multicolumn{1}{c|}{69.70}          & 67.10                     & \multicolumn{1}{c|}{62.30}          & 63.50                     & 63.50          \\
DialogueGCN\cite{DialogueGCN}              & 40.62                     & \multicolumn{1}{c|}{42.75}          & 89.14                     & \multicolumn{1}{c|}{84.54}          & 61.92                     & \multicolumn{1}{c|}{63.54}          & 67.53                     & \multicolumn{1}{c|}{64.19}          & 65.46                     & \multicolumn{1}{c|}{63.08}          & 64.18                     & \multicolumn{1}{c|}{66.99}          & 65.25                     & 64.18          \\ \hline
DialogueGCN(80:20)              & 52.83                     & \multicolumn{1}{c|}{42.47}          & 79.43                     & \multicolumn{1}{c|}{77.26}          & 60.93                     & \multicolumn{1}{c|}{58.48}          & 61.89                     & \multicolumn{1}{c|}{57.82}          & 66.85                     & \multicolumn{1}{c|}{\textbf{74.91}} & 56.28                     & \multicolumn{1}{c|}{56.82}          & 63.23                     & 62.46          \\
BERT\cite{BERT}                     & 42.19                     & \multicolumn{1}{c|}{39.05}          & 60.45                     & \multicolumn{1}{c|}{59.91}          & 49.11                     & \multicolumn{1}{c|}{52.24}          & 55.14                     & \multicolumn{1}{c|}{54.82}          & 64.22                     & \multicolumn{1}{c|}{55.97}          & 54.26                     & \multicolumn{1}{c|}{55.88}          & 54.06                     & 54.01          \\ \hline
TRM                      & 43.08                     & \multicolumn{1}{c|}{42.42}          & 77.27                     & \multicolumn{1}{c|}{74.54}          & 58.24                     & \multicolumn{1}{c|}{60.88}          & 65.00                     & \multicolumn{1}{c|}{60.22}          & 68.37                     & \multicolumn{1}{c|}{66.98}          & 61.48                     & \multicolumn{1}{c|}{59.84}          & 62.25                     & 62.11          \\
TRMSM-Add                & 43.53                     & \multicolumn{1}{c|}{48.53}          & 76.15                     & \multicolumn{1}{c|}{76.74}          & 66.06                     & \multicolumn{1}{c|}{\textbf{64.37}} & 56.24                     & \multicolumn{1}{c|}{60.6}           & 75.72                     & \multicolumn{1}{c|}{68.49}          & 63.23                     & \multicolumn{1}{c|}{61.18}          & 64.21                     & 64.45          \\
TRMSM-Cat                & 48.71                     & \multicolumn{1}{c|}{49.46}          & 76.84                     & \multicolumn{1}{c|}{77.02}          & 64.44                     & \multicolumn{1}{c|}{63.00}          & 57.03                     & \multicolumn{1}{c|}{60.72}          & 76.07                     & \multicolumn{1}{c|}{70.33}          & 60.74                     & \multicolumn{1}{c|}{62.09}          & 64.72                     & 64.82          \\
TRMSM-Att                & 43.36                     & \multicolumn{1}{c|}{50.22} & 81.23                     & \multicolumn{1}{c|}{75.82}          & 66.11                     & \multicolumn{1}{c|}{64.15}          & 60.39                     & \multicolumn{1}{c|}{60.97}          & 77.46                     & \multicolumn{1}{c|}{72.70}          & 62.16                     & \multicolumn{1}{c|}{\textbf{63.45}} & 65.34                     & \textbf{65.74} \\ \bottomrule
\end{tabular}}
\caption{The results of our models on IEMOCAP. Weighted-F1 score (wF1) is used as the metric. \dag  \ means referring from \citet{DialogueGCN}. }
\label{tab: iem}
\end{table*}

\section{Experimental Setup}

\subsection{Datasets}
We evaluate our models on two datasets: IEMOCAP \citep{IEMOCAP}, MELD \cite{MELD}, and both of them are multi-modal datasets that contain three modalities. We solely consider the textual modality following \citet{DialogueGCN}. Statistics about the datasets are shown in Tab. \ref{tab:dataset}. 
\begin{itemize}
    \item \textbf{IEMOCAP}\quad This dataset contains a series of dyadic conversation between 10 unique speakers. 6 categories of emotions are considered in our experiments: \textit{neutral}, \textit{happy}, \textit{sad}, \textit{angry}, \textit{excited}, and \textit{frustrated}. Following \citet{DialogueRNN}, the training set is split into a new training set and a validation set by the ratio of 80: 20. 
    
   \item \textbf{MELD}\quad This dataset contains over 1400 multi-speakers conversations collected from the TV series \textit{Friends}.  Emotions in this dataset are annotated into 7 categories: \textit{neutral}, \textit{joy}, \textit{surprise}, \textit{anger}, \textit{disgust}, \textit{sadness} and \textit{fear}.
\end{itemize}

\subsection{Compared Methods}

To distinguish our models with different fusing methods, we construct 3 model variants: TRMSM-Add, TRMSM-Cat, and TRMSM-Att. To show the importance of the speaker-related information, we construct our model without Intra-Speaker Blocks and Inter-Speaker Blocks, which we denote it as TRM. Besides, our models are compared with the baselines below:

\begin{itemize}
    \item \textbf{CMN \citep{CMN}}\quad CMN is proposed to model dyadic conversations using two sets of RNNs and Memnets to respectively track different speakers. 
    \item \textbf{DialogueRNN \citep{DialogueRNN}}\quad DialogueRNN is the first state-of-the-art model to address ERC of multi-speakers. RNNs are deployed to track speakers' states and global state during conversations. 
    \item \textbf{AGHMN \citep{AGHMN}}\quad AGHMN is the state-of-the-art unidirectional model in real-time ERC. To retain the positional information from its hierarchical structure, a GRU is constructed for attention mechanism. 
    \item \textbf{DialogueGCN \citep{DialogueGCN}}\quad  To fully model the interactive information between speakers, DialogueGCN models detailed dependencies between speakers using a Relational GCN. 
    \item \textbf{KET \citep{KET}}\quad  KET introduces the Transformer structure to model context in conversations. It also proposes an effective graph attention to extract information from commonsense knowledge bases. 
    \item \textbf{BERT \citep{BERT}}\quad  A vanilla BERT followed by a classifier is fine-tuned to show the importance of context. 
    \item \textbf{Other baselines}\quad Both based on CNN to extract semantic information, scLSTM \citep{bcLSTM} utilizes LSTM \citep{LSTM} and Memnet \citep{MN} utilizes memory network to model conversational context. 
\end{itemize}

\subsection{Implementation}

For BERT and sentence-level encoder, an uncased BERT-base\footnote{https://github.com/huggingface/transformers} model is adopted. For the dialogue-level encoder, the dimension of dialogue-level representation is set to 300 for IEMOCAP and 200 for MELD; the number of transformer layers is set to 6 for IEMOCAP and 1 for MELD; the number of heads is set 6 for IEMOCAP and 4 for MELD; dropout rate is set to $0.1$. Additionally, models are trained using AdamW \citep{Adam,Weight} for 10000 steps with 1000 steps for warming up, and the learning rate linearly decaying after the warm-up is set to 1e-5 for IEMOCAP and 8e-6 for MELD. Due to the parallel prediction of utterances in one conversation, batch size is set to 1 following \citet{AGHMN}. Besides, DialogueGCN is trained in the setting of 90:10 data split on IEMOCAP, and for a fair comparison, we re-run DialogueGCN with 80:20 data split using the open-source code\footnote{https://github.com/declare-lab/conv-emotion/tree/master/DialogueGCN}. All of our results reported are the average values of 5 runs.

\begin{table*}
\centering
\scalebox{0.67}{
\begin{tabular}{@{}l|ccccccccccccccccc@{}}
\toprule
\multirow{3}{*}{Methods} & \multicolumn{17}{c}{MELD}                                                                                                                                                                                                                                                                                                                                                                                                                                                                                                                        \\ \cline{2-18} 
                         & \multicolumn{2}{c|}{Neutral}                                    & \multicolumn{2}{c|}{Surprise}                                   & \multicolumn{2}{c|}{Fear}                                       & \multicolumn{2}{c|}{Sadness}                                    & \multicolumn{2}{c|}{Joy}                                        & \multicolumn{2}{c|}{Disgust}                                    & \multicolumn{2}{c|}{Anger}                                      & \multicolumn{3}{c}{Average}                                        \\ \cline{2-18} 
                         & \multicolumn{1}{c|}{Acc.} & \multicolumn{1}{c|}{F1}             & \multicolumn{1}{c|}{Acc.} & \multicolumn{1}{c|}{F1}             & \multicolumn{1}{c|}{Acc.} & \multicolumn{1}{c|}{F1}             & \multicolumn{1}{c|}{Acc.} & \multicolumn{1}{c|}{F1}             & \multicolumn{1}{c|}{Acc.} & \multicolumn{1}{c|}{F1}             & \multicolumn{1}{c|}{Acc.} & \multicolumn{1}{c|}{F1}             & \multicolumn{1}{c|}{Acc.} & \multicolumn{1}{c|}{wF1}            & 
                         \multicolumn{1}{c|}{Acc.} & \multicolumn{1}{c|}{wF1}            & \multicolumn{1}{c}{mF1}  \\ \hline\hline
scLSTM\cite{bcLSTM}$\diamondsuit$                   & 78.40                     & \multicolumn{1}{c|}{73.80}          & 46.80                     & \multicolumn{1}{c|}{47.70}          & 3.80                      & \multicolumn{1}{c|}{5.40}           & 22.40                     & \multicolumn{1}{c|}{25.10}          & 51.60                     & \multicolumn{1}{c|}{51.30}          & 4.30                      & \multicolumn{1}{c|}{5.20}           & 36.70                     & \multicolumn{1}{c|}{38.40}          & 57.50                 & 55.90                     & 46.40          \\
DialogueRNN\cite{DialogueRNN}$\diamondsuit$              & 72.10                     & \multicolumn{1}{c|}{73.5}           & 54.40                     & \multicolumn{1}{c|}{49.40}          & 1.60                      & \multicolumn{1}{c|}{1.20}           & 23.90                     & \multicolumn{1}{c|}{23.80}          & 52.00                     & \multicolumn{1}{c|}{50.70}          & 1.50                      & \multicolumn{1}{c|}{1.70}           & 41.90                     & \multicolumn{1}{c|}{41.50}          & 56.10                 & 55.90                     & 45.30          \\
AGHMN\cite{AGHMN}                    & 83.40                     & \multicolumn{1}{c|}{76.40}          & 49.10                     & \multicolumn{1}{c|}{49.70}          & 9.20                      & \multicolumn{1}{c|}{11.50}          & 21.60                     & \multicolumn{1}{c|}{27.00}          & 52.40                     & \multicolumn{1}{c|}{52.40}          & 12.20                     & \multicolumn{1}{c|}{14.00}          & 34.90                     & \multicolumn{1}{c|}{39.40}          & 60.30                 & 58.10                     & 49.45          \\
KET\cite{KET}                      & -                         & \multicolumn{1}{c|}{-}              & -                         & \multicolumn{1}{c|}{-}              & -                         & \multicolumn{1}{c|}{-}              & -                         & \multicolumn{1}{c|}{-}              & -                         & \multicolumn{1}{c|}{-}              & -                         & \multicolumn{1}{c|}{-}              & -                         & \multicolumn{1}{c|}{-}              & -                     & 58.18                     & -              \\
DialogueGCN\cite{DialogueGCN}              & -                         & \multicolumn{1}{c|}{-}              & -                         & \multicolumn{1}{c|}{-}              & -                         & \multicolumn{1}{c|}{-}              & -                         & \multicolumn{1}{c|}{-}              & -                         & \multicolumn{1}{c|}{-}              & -                         & \multicolumn{1}{c|}{-}              & -                         & \multicolumn{1}{c|}{-}              & \multicolumn{1}{l}{-} & \multicolumn{1}{l}{58.10} & -              \\ \hline
BERT\cite{BERT}                     & 74.85                     & \multicolumn{1}{c|}{76.57}          & 53.26                     & \multicolumn{1}{c|}{56.25}          & 21.94                     & \multicolumn{1}{c|}{21.89}          & 35.33                     & \multicolumn{1}{c|}{33.01}          & 53.82                     & \multicolumn{1}{c|}{57.02}          & 36.64                     & \multicolumn{1}{c|}{27.67} & 50.70                     & \multicolumn{1}{c|}{42.42}          & 61.82                 & 61.07                     & 53.40          \\ \hline
TRM                      & 76.57                     & \multicolumn{1}{c|}{76.62}          & 55.53                     & \multicolumn{1}{c|}{55.67}          & 23.9                      & \multicolumn{1}{c|}{\textbf{23.8}}           & 36.43                     & \multicolumn{1}{c|}{32.34}          & 52.07                     & \multicolumn{1}{c|}{57.46}          & 29.46                     & \multicolumn{1}{c|}{25.13}          & 50.68                     & \multicolumn{1}{c|}{44.64}          & 61.80                 & 61.30                     & 53.45          \\
TRMSM-Add                & 75.88                     & \multicolumn{1}{c|}{\textbf{77.71}}          & 53.40                     & \multicolumn{1}{c|}{56.49}          & 24.67                     & \multicolumn{1}{c|}{21.14}          & 36.39                     & \multicolumn{1}{c|}{31.56} & 53.97                     & \multicolumn{1}{c|}{57.83}          & 35.07                     & \multicolumn{1}{c|}{22.62}          & 52.36                     & \multicolumn{1}{c|}{45.95}          & 62.93                 & 62.06                     & 53.96          \\
TRMSM-Cat                & 75.69                     & \multicolumn{1}{c|}{77.26}          & 52.64                     & \multicolumn{1}{c|}{56.33}          & 25.66                     & \multicolumn{1}{c|}{22.73}          & 38.41                     & \multicolumn{1}{c|}{\textbf{34.37}}          & 57.81                     & \multicolumn{1}{c|}{58.07}          & 35.61                     & \multicolumn{1}{c|}{22.57}          & 48.34                     & \multicolumn{1}{c|}{45.90}          & 62.76                 & 62.01                     & 54.04          \\
TRMSM-Att                & 75.48                     & \multicolumn{1}{c|}{77.56} & 55.90                     & \multicolumn{1}{c|}{\textbf{57.25}} & 25.91                     & \multicolumn{1}{c|}{20.38} & 36.82                     & \multicolumn{1}{c|}{32.9}          & 55.55                     & \multicolumn{1}{c|}{\textbf{58.66}} & 38.31                     & \multicolumn{1}{c|}{\textbf{28.63}}          & 52.11                     & \multicolumn{1}{c|}{\textbf{45.95}} & 63.23                 & \textbf{62.36}            & \textbf{54.57} \\ \bottomrule
\end{tabular}
}
\caption{The Results of our models on MELD. MELD uses weighted-F1 (wF1) score, and the average value (mF1) of Macro-F1 and Micro-F1, as the metrics. $\diamondsuit$ means referring from \citet{AGHMN}. }
\label{tab: meld}
\end{table*}

\section{Results and Discussions}

\subsection{Overall Results}

For IEMOCAP, weighted-F1 (wF1) score is used as the metric. However, the data proportion of MELD is in a severely imbalanced condition. Therefore, the weighted-F1 score is not that proper and enough for MELD. To balance the contributions of large classes and small classes, we follow \citet{Stance} and also use the average value of macro F1 score and micro F1 score as one metric, which is calculated by $mF1=(F1_{macro}+F1_{micro})/2$. 

\begin{figure}
    \centering
    \includegraphics[width=0.47\textwidth]{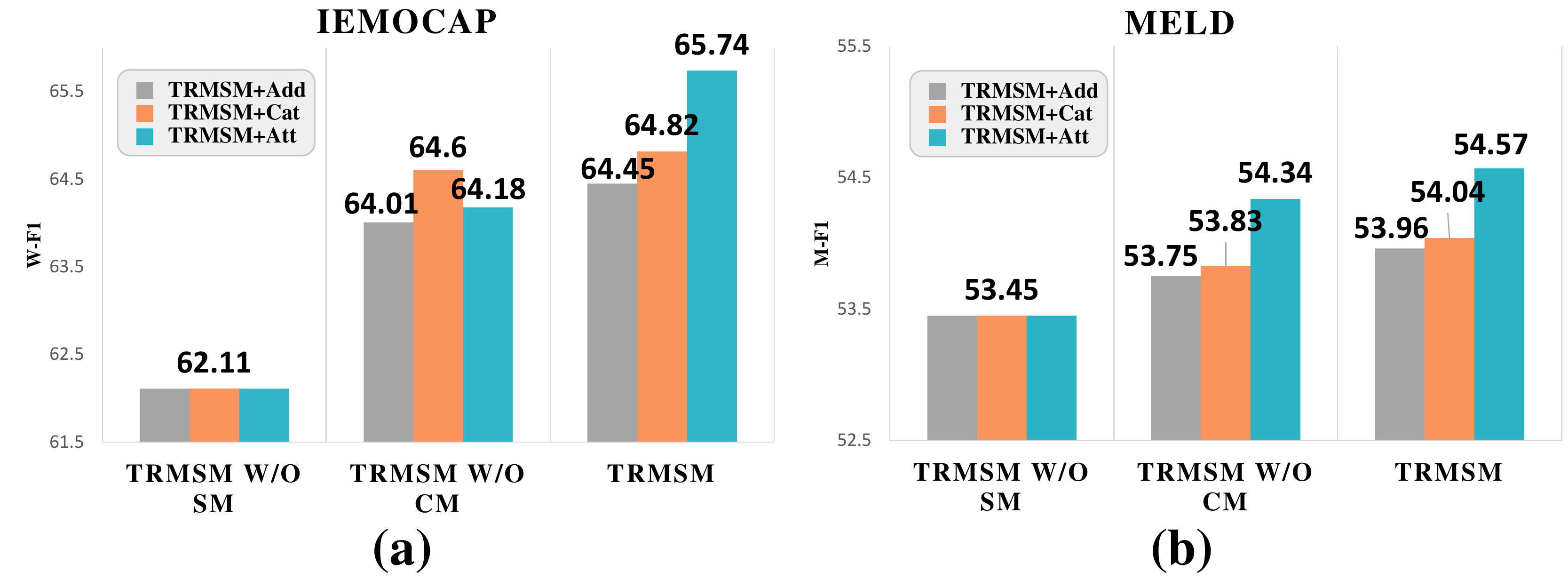}
    \caption{The results of models with different blocks. wF1 for IEMOCAP; mF1 for MELD.}
    \label{fig: ablation}
\end{figure}

For IEMOCAP, as shown in Tab. \ref{tab: iem}, BERT attains wF1 of 54.01 which is substantially worse than our models and most state-of-the-art models considering the dialogue-level context. This result may indicate that IEMOCAP contains considerable utterances that cannot be predicted only depending on the semantic information, which is out of the conversational context. Furthermore, TRMSM-Att outperforms AGHMN by 2.24 wF1 and DialogueGCN (80: 20) by 3.28 wF1, which benefits from the powerful Transformer and our speaker modeling with long-distance information considered. For emotions, TRMSM-Att achieves the best F1 on \textit{Frustrated}, and TRMSM-Add achieves the best F1 on \textit{Neutral}. Besides, our models can attain second or third higher results among other emotions. This demonstrates that our models are competitive to achieve comprehensive performance. 

For MELD, as shown in Tab. \ref{tab: meld}, BERT outperforms other state-of-the-art models by a great margin, which indicates the importance of external knowledge brought by BERT.  Compared with BERT, TRM attains marginally better results, which may be attributed to the limited conversational contextual information in MELD. To confirm this, comparing the results of CNN\footnote{CNN achieves 55.02 wF1 on MELD by \citet{Survey}.} and scLSTM (based on CNN), we can notice that the improvement is also limited. Although MELD provides limited contextual information in conversations, TRMSM-Att still outperforms BERT by 1.29 wF1 and 1.17 mF1, which indicates the effectiveness of our model to capture such information. For emotions, BERT beats other state-of-the-art models by a great margin in such an imbalanced circumstance, and TRMSM-Att attains the best F1 on 4 emotions including large classes \textit{Joy}, \textit{Anger}, \textit{Surprise}, and the small class \textit{Disgust}. This demonstrates that BERT can alleviate data imbalance and our model can take advantage of such a feature. 

For both datasets, all TRMSM variants outperform TRM to show the importance of speaker-aware contextual information. TRMSM-Att and TRMSM-Cat outperform TRMSM-Add, which indicates the importance of different aspects of speaker information requiring to be treated differently. TRMSM-Att outperforming TRMSM-Cat demonstrates that automatically and explicitly picking up speaker-related information is better than the implicit way. 

\begin{figure}
    \centering
    \includegraphics[width=0.47\textwidth]{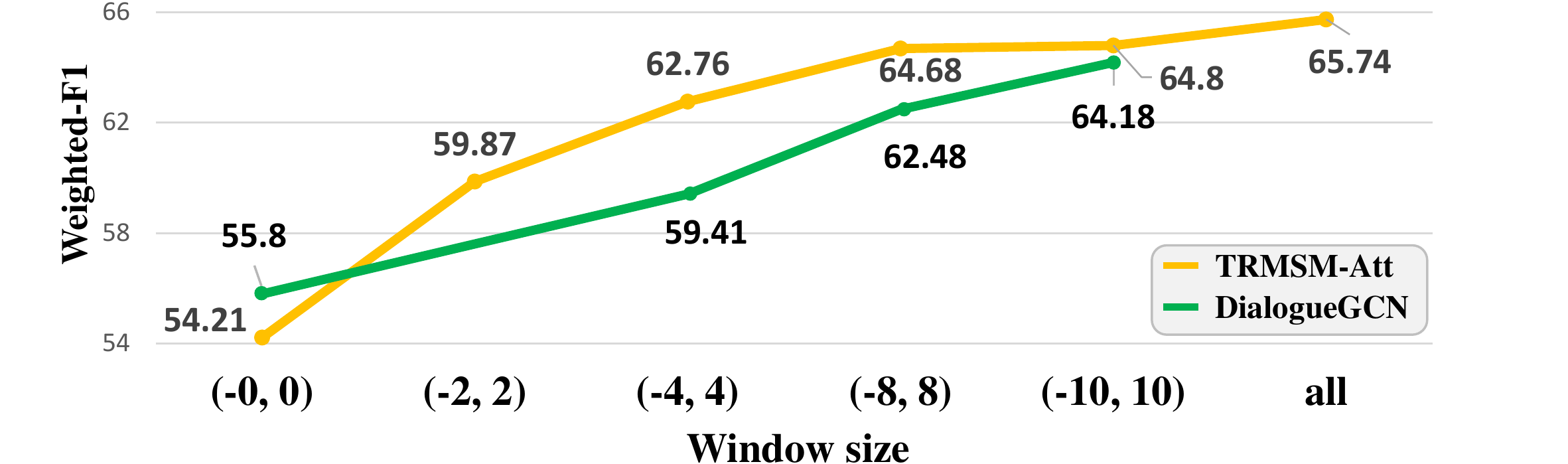}
    \caption{wF1 of TRMSM-Att and DialogueGCN with different ranges of context. \textit{all} means using global context. }
    \label{fig: win}
\end{figure}

\subsection{Model Analysis}

\begin{figure*}
    \centering
    \includegraphics[width=1\textwidth]{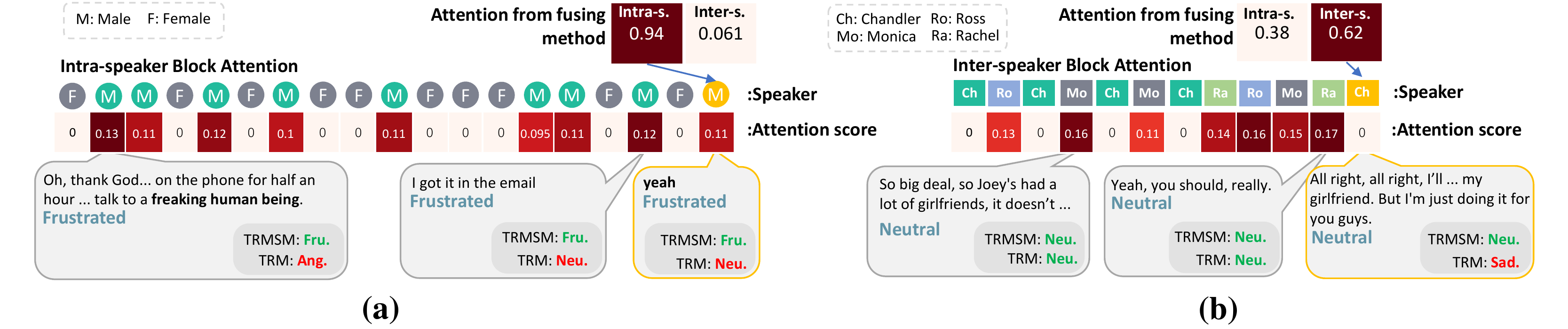}
    \caption{Heatmaps of attention from fusing method and self-attention of Intra-, Inter-Speaker Blocks for the targeted utterances (whose speakers are marked in yellow). Labels of utterances are tagged below the utterances. Predictions of TRMSM and TRM are marked in green for correctness and red for mistake. }
    \label{fig: case}
\end{figure*}

\begin{figure}
    \centering
    \includegraphics[width=0.47\textwidth]{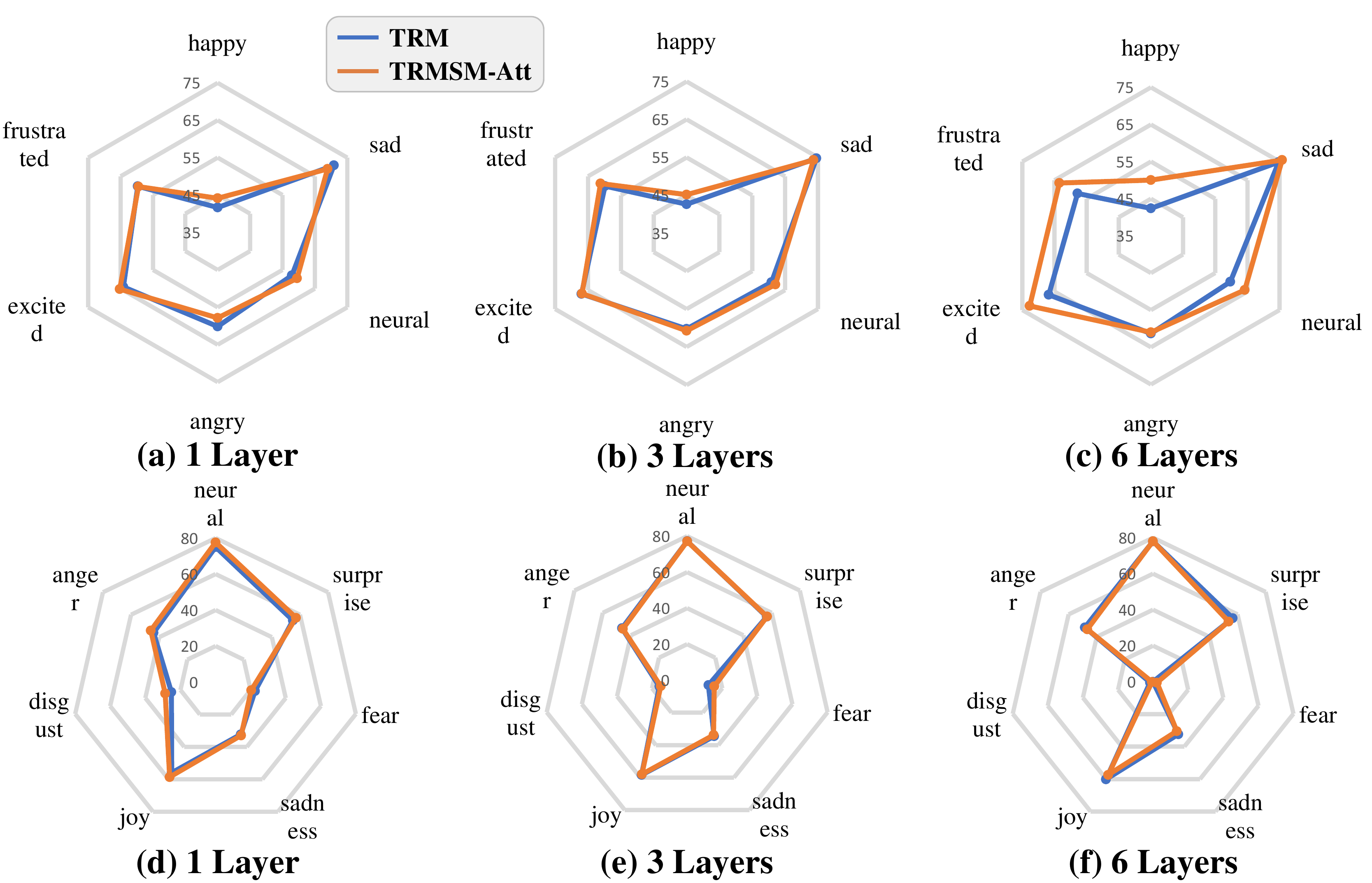}
    \caption{The F1 score on every emotion class by TRMSM-Att and TRM. (a)-(c) for IEMOCAP; (d)-(f) for MELD.}
    \label{fig: layer}
\end{figure}

\quad \textbf{Ablation Study} \quad To better understand the influences of masks on our models, we report the results of the models removing the Transformer blocks with different masks on IEMOCAP and MELD. In this part, we denote Convention Mask as CM and Intra-Speaker, Inter-Speaker Masks as SM. Accordingly, TRMSM w/o SM is equivalent to TRM. 

As seen in Fig. \ref{fig: ablation}, on both datasets, TRMSM w/o CM (solely applying SM) can achieve better performance than TRMSM w/o SM (solely applying CM). We attribute it to that speaker modeling does not drop the contextual information from conversations, and on the contrary, speaker modeling can guide the model to extract more effective information to the final prediction. Furthermore, TRMSM outperforms both TRMSM w/o CM and TRMSM w/o SM, which demonstrates that all of our designed masks are critical to achieving better performance. 


\textbf{Effect of Range of Context} \quad To find out the influence of the range of context on our model, we train TRMSM-Att with different ranges of available context on IEMOCAP and refer the results from \citet{DialogueGCN} for DialogueGCN. We utilize different windows $(-x, y)$ to limit the context, where $x$, $y$ is respectively the number of utterances in prior context and post context. As illustrated in Fig. \ref{fig: win}, with the window widened, the performance increases as shown in both models. For DialogueGCN, (-10, 10) is the max window of context and therefore it cannot get access to the long-distance context. On the contrary, the performance is further improved by TRMSM-Att with all context available. This indicates that the local contextual information is critical for the prediction and the long-distance information is also important for contextual modeling to further improve the performance. 

\textbf{Effect of Number of Layers} \quad We study the effect of the number of layers to our model on different datasets. Fig. \ref{fig: layer} illustrates the radar graphs for the F1 scores of emotions in IEMOCAP and MELD by TRM and TRMSM-Att. As the number of layers increasing, the F1 scores of emotions in IEMOCAP normally expand. While in MELD, increasing the number of layers gradually hurts the performance to be 0 of F1 on emotions \textit{Fear} and \textit{Disgust} which are classes with the fewest data. We think the reason may be that MELD suffers from data imbalance and increasing the number of layers leads to severer overfitting on small classes. For data imbalance, methods like re-balance can be applied to alleviate it. Re-balance is out of the scope of this paper and our future work will study data imbalance of ERC. 

\subsection{Case Study}

To better understand how our model captures \emph{Intra-Speaker} and \emph{Inter-Speaker} dependencies, we illustrate two conversation clips ending with the targeted utterances so that the targeted utterances can only refer to the prior context. We choose TRMSM-Att without Conventional Blocks so that only speaker information related blocks are considered. Specifically, we illustrate heatmaps of attention from fusing method and self-attention\footnote{We average the attention scores of all self-attention heads in the top layer of Transformer} in Transformer blocks. For simplicity, we denote attention from fusing method as FAtt. 

In the scene of Fig. \ref{fig: case} (a), the speaker M keeps in frustration through the conversation and F in a neutral state has few influences on M. Therefore, FAtt pays more attention to \emph{Intra-Speaker} dependency so that Intra-Speaker Blocks can extract information from M himself. We can see from the heatmap that the targeted utterance \textit{yeah} grades the highest score to the farthest contextual utterance whose emotion is also frustration, which is out of the range of context that DialogueGCN can refer. In a sense, this indicates the importance of long-distance information. 

As the condition in Fig. \ref{fig: case} (b), speakers in this conversation basically keep in a neutral state except that Chandler shows other emotions like anger and surprise before the targeted utterance. Although the targeted utterance with speaker Chandler shows slight sadness from the semantic view, it is supposed to be predicted as neutral according to the conversational context. Specifically, FAtt grades Inter-Speaker Blocks with higher score and self-attention in Inter-Speaker Blocks extracts information from the neutral utterances of other speakers. This case indicates the effectiveness of our model to extract inter-speaker information. 

\section{Conclusion}
In this work, we simplify the Self and Inter-Speaker dependencies to a binary version. To achieve the simplified modeling of speakers' interactions, we design three masks: Conventional Mask, Intra-Speaker Mask, and Inter-Speaker Mask. These masks are utilized in the self-attention modules of the second-level Transformer blocks of a hierarchical Transformer. As the speaker-aware information extracted by different masks diversely contributes to the prediction, attention mechanism is utilized to weight and fuse them. Finally, our model achieves state-of-the-art results on 2 ERC datasets and further analysis shows that our model is efficacious for ERC. 

\bibliography{aaai21}

\begin{thebibliography}{25}
\providecommand{\natexlab}[1]{#1}
\providecommand{\url}[1]{\texttt{#1}}
\providecommand{\urlprefix}{URL }
\expandafter\ifx\csname urlstyle\endcsname\relax
  \providecommand{\doi}[1]{doi:\discretionary{}{}{}#1}\else
  \providecommand{\doi}{doi:\discretionary{}{}{}\begingroup
  \urlstyle{rm}\Url}\fi

\bibitem[{Busso et~al.(2008)Busso, Bulut, Lee, Kazemzadeh, Mower, Kim, Chang,
  Lee, and Narayanan}]{IEMOCAP}
Busso, C.; Bulut, M.; Lee, C.; Kazemzadeh, A.; Mower, E.; Kim, S.; Chang,
  J.~N.; Lee, S.; and Narayanan, S.~S. 2008.
\newblock {IEMOCAP:} interactive emotional dyadic motion capture database.
\newblock \emph{Lang. Resour. Evaluation} .

\bibitem[{Devlin et~al.(2019)Devlin, Chang, Lee, and Toutanova}]{BERT}
Devlin, J.; Chang, M.; Lee, K.; and Toutanova, K. 2019.
\newblock {BERT:} Pre-training of Deep Bidirectional Transformers for Language
  Understanding.
\newblock In \emph{Proc. of NAACL-HLT}, 4171--4186.

\bibitem[{Dong et~al.(2019)Dong, Yang, Wang, Wei, Liu, Wang, Gao, Zhou, and
  Hon}]{UniLM}
Dong, L.; Yang, N.; Wang, W.; Wei, F.; Liu, X.; Wang, Y.; Gao, J.; Zhou, M.;
  and Hon, H. 2019.
\newblock Unified Language Model Pre-training for Natural Language
  Understanding and Generation.
\newblock In \emph{Proc. of NeurIPS}, 13042--13054.

\bibitem[{Ghosal et~al.(2019)Ghosal, Majumder, Poria, Chhaya, and
  Gelbukh}]{DialogueGCN}
Ghosal, D.; Majumder, N.; Poria, S.; Chhaya, N.; and Gelbukh, A.~F. 2019.
\newblock DialogueGCN: {A} Graph Convolutional Neural Network for Emotion
  Recognition in Conversation.
\newblock In \emph{Proc. of EMNLP-IJCNLP}, 154--164.

\bibitem[{Hazarika et~al.(2018{\natexlab{a}})Hazarika, Poria, Mihalcea,
  Cambria, and Zimmermann}]{ICON}
Hazarika, D.; Poria, S.; Mihalcea, R.; Cambria, E.; and Zimmermann, R.
  2018{\natexlab{a}}.
\newblock {ICON:} Interactive Conversational Memory Network for Multimodal
  Emotion Detection.
\newblock In \emph{Proc. of EMNLP}, 2594--2604.

\bibitem[{Hazarika et~al.(2018{\natexlab{b}})Hazarika, Poria, Zadeh, Cambria,
  Morency, and Zimmermann}]{CMN}
Hazarika, D.; Poria, S.; Zadeh, A.; Cambria, E.; Morency, L.; and Zimmermann,
  R. 2018{\natexlab{b}}.
\newblock Conversational Memory Network for Emotion Recognition in Dyadic
  Dialogue Videos.
\newblock In \emph{Proc. of NAACL-HLT}, 2122--2132.

\bibitem[{Hochreiter and Schmidhuber(1997)}]{LSTM}
Hochreiter, S.; and Schmidhuber, J. 1997.
\newblock Long Short-Term Memory.
\newblock \emph{Neural Computation} 9(8): 1735--1780.

\bibitem[{Jiao, Lyu, and King(2020)}]{AGHMN}
Jiao, W.; Lyu, M.~R.; and King, I. 2020.
\newblock Real-Time Emotion Recognition via Attention Gated Hierarchical Memory
  Network.
\newblock In \emph{Proc. of AAAI}.

\bibitem[{Jiao et~al.(2019)Jiao, Yang, King, and Lyu}]{HiGRUs}
Jiao, W.; Yang, H.; King, I.; and Lyu, M.~R. 2019.
\newblock HiGRU: Hierarchical Gated Recurrent Units for Utterance-Level Emotion
  Recognition.
\newblock In \emph{Proc. of NAACL-HLT}, 397--406.

\bibitem[{Kingma and Ba(2015)}]{Adam}
Kingma, D.~P.; and Ba, J. 2015.
\newblock Adam: {A} Method for Stochastic Optimization.
\newblock In \emph{Proc. of ICLR}.

\bibitem[{Li et~al.(2020)Li, Wu, Zheng, and Wang}]{HiTrans}
Li, Q.; Wu, C.; Zheng, K.; and Wang, Z. 2020.
\newblock Hierarchical Transformer Network for Utterance-level Emotion
  Recognition.
\newblock \emph{arXiv preprint arXiv:2002.07551} .

\bibitem[{Lian et~al.(2019)Lian, Tao, Liu, and Huang}]{Fusion}
Lian, Z.; Tao, J.; Liu, B.; and Huang, J. 2019.
\newblock Conversational Emotion Analysis via Attention Mechanisms.
\newblock In \emph{Proc. of Interspeech}, 1936--1940.

\bibitem[{Loshchilov and Hutter(2019)}]{Weight}
Loshchilov, I.; and Hutter, F. 2019.
\newblock Decoupled Weight Decay Regularization.
\newblock In \emph{Proc. of ICLR}.

\bibitem[{Majumder et~al.(2019)Majumder, Poria, Hazarika, Mihalcea, Gelbukh,
  and Cambria}]{DialogueRNN}
Majumder, N.; Poria, S.; Hazarika, D.; Mihalcea, R.; Gelbukh, A.~F.; and
  Cambria, E. 2019.
\newblock DialogueRNN: An Attentive {RNN} for Emotion Detection in
  Conversations.
\newblock In \emph{Proc. of AAAI}, 6818--6825.

\bibitem[{Poria et~al.(2017)Poria, Cambria, Hazarika, Majumder, Zadeh, and
  Morency}]{bcLSTM}
Poria, S.; Cambria, E.; Hazarika, D.; Majumder, N.; Zadeh, A.; and Morency, L.
  2017.
\newblock Context-Dependent Sentiment Analysis in User-Generated Videos.
\newblock In \emph{Proc. of ACL}, 873--883.

\bibitem[{Poria et~al.(2019{\natexlab{a}})Poria, Hazarika, Majumder, Naik,
  Cambria, and Mihalcea}]{MELD}
Poria, S.; Hazarika, D.; Majumder, N.; Naik, G.; Cambria, E.; and Mihalcea, R.
  2019{\natexlab{a}}.
\newblock {MELD:} {A} Multimodal Multi-Party Dataset for Emotion Recognition in
  Conversations.
\newblock In \emph{Proc. of ACL}, 527--536.

\bibitem[{Poria et~al.(2019{\natexlab{b}})Poria, Majumder, Mihalcea, and
  Hovy}]{Survey}
Poria, S.; Majumder, N.; Mihalcea, R.; and Hovy, E.~H. 2019{\natexlab{b}}.
\newblock Emotion Recognition in Conversation: Research Challenges, Datasets,
  and Recent Advances.
\newblock \emph{{IEEE} Access} 7: 100943--100953.

\bibitem[{Radford et~al.(2018)Radford, Narasimhan, Salimans, and
  Sutskever}]{GPT}
Radford, A.; Narasimhan, K.; Salimans, T.; and Sutskever, I. 2018.
\newblock Improving language understanding by generative pre-training.

\bibitem[{Schlichtkrull et~al.(2018)Schlichtkrull, Kipf, Bloem, van~den Berg,
  Titov, and Welling}]{RGCN}
Schlichtkrull, M.~S.; Kipf, T.~N.; Bloem, P.; van~den Berg, R.; Titov, I.; and
  Welling, M. 2018.
\newblock Modeling Relational Data with Graph Convolutional Networks.
\newblock In \emph{Proc. of ESWC}, 593--607.

\bibitem[{Sukhbaatar et~al.(2015)Sukhbaatar, Szlam, Weston, and Fergus}]{MN}
Sukhbaatar, S.; Szlam, A.; Weston, J.; and Fergus, R. 2015.
\newblock End-To-End Memory Networks.
\newblock In \emph{Proc. of NeuIPS}, 2440--2448.

\bibitem[{Vaswani et~al.(2017)Vaswani, Shazeer, Parmar, Uszkoreit, Jones,
  Gomez, Kaiser, and Polosukhin}]{Transformer}
Vaswani, A.; Shazeer, N.; Parmar, N.; Uszkoreit, J.; Jones, L.; Gomez, A.~N.;
  Kaiser, L.; and Polosukhin, I. 2017.
\newblock Attention is All you Need.
\newblock In \emph{Proc. of NeuIPS}, 5998--6008.

\bibitem[{Zhang et~al.(2020)Zhang, Yang, Li, Ye, Xu, and Dai}]{Stance}
Zhang, B.; Yang, M.; Li, X.; Ye, Y.; Xu, X.; and Dai, K. 2020.
\newblock Enhancing Cross-target Stance Detection with Transferable
  Semantic-Emotion Knowledge.
\newblock In \emph{Proc. of ACL}, 3188--3197.

\bibitem[{Zhang et~al.(2019)Zhang, Wu, Sun, Li, Zhu, and Zhou}]{ConvGCN}
Zhang, D.; Wu, L.; Sun, C.; Li, S.; Zhu, Q.; and Zhou, G. 2019.
\newblock Modeling both Context- and Speaker-Sensitive Dependence for Emotion
  Detection in Multi-speaker Conversations.
\newblock In \emph{Proc. of IJCAI}, 5415--5421.

\bibitem[{Zhong, Wang, and Miao(2019)}]{KET}
Zhong, P.; Wang, D.; and Miao, C. 2019.
\newblock Knowledge-Enriched Transformer for Emotion Detection in Textual
  Conversations.
\newblock In \emph{Proc. of EMNLP-IJCNLP}, 165--176.

\bibitem[{Zhu et~al.(2020)Zhu, Nan, Wang, Nallapati, and Xiang}]{StrucureTrans}
Zhu, H.; Nan, F.; Wang, Z.; Nallapati, R.; and Xiang, B. 2020.
\newblock Who did They Respond to? Conversation Structure Modeling using Masked
  Hierarchical Transformer.
\newblock In \emph{Proc. of AAAI}.

\end{thebibliography}
\end{document}